\pdfoutput=1
\documentclass[10pt,twocolumn,letterpaper]{article}
\usepackage{wacv}
\usepackage{footnote}
\usepackage{times}
\usepackage{epsfig}
\usepackage{graphicx}
\usepackage{amsmath}
\usepackage{amssymb}
\usepackage{epstopdf}
\usepackage{array}
\usepackage{subfigure}
\usepackage{multirow} 
\usepackage{caption}
\usepackage{mathtools}

\wacvfinalcopy 

\def\wacvPaperID{371} 

\ifwacvfinal\fi
\begin{document}

\title{Interactively Test Driving an Object Detector: Estimating Performance on Unlabeled Data}

\author{Rushil Anirudh and Pavan Turaga\\
School of Electrical, Computer and Energy Engineering\\
School of Arts, Media and Engineering\\
Arizona State University, Tempe.\\
{\tt\small \{ranirudh,pturaga\}@asu.edu}
}
\maketitle
 \thispagestyle{empty}
\begin{abstract}
In this paper, we study the problem of `test-driving' a detector, i.e. allowing a human user to get a quick sense of how well the detector generalizes to their specific requirement. To this end, we present the first system that estimates detector performance interactively without extensive ground truthing using a human in the loop. We approach this as a problem of estimating proportions and show that it is possible to make accurate inferences on the proportion of classes or groups within a large data collection by observing only $5-10\%$ of samples from the data. In estimating the false detections (for precision), the samples are chosen carefully such that the overall characteristics of the data collection are preserved. Next, inspired by its use in estimating disease propagation we apply pooled testing approaches to estimate missed detections (for recall) from the dataset. The estimates thus obtained are close to the ones obtained using ground truth, thus reducing the need for extensive labeling which is expensive and time consuming.\let\thefootnote\relax\footnotetext{This work was supported in part by the NSF-CCF-CIF award 1320267, ONR Grant N00014-12-1-0124 subaward Z868302 and by ASU startup grant.}

\end{abstract}

\section{Introduction}
\begin{figure}[!htb]
\centering
\includegraphics[trim = 65mm 25mm 45mm 0mm, clip,width = 8.5cm]{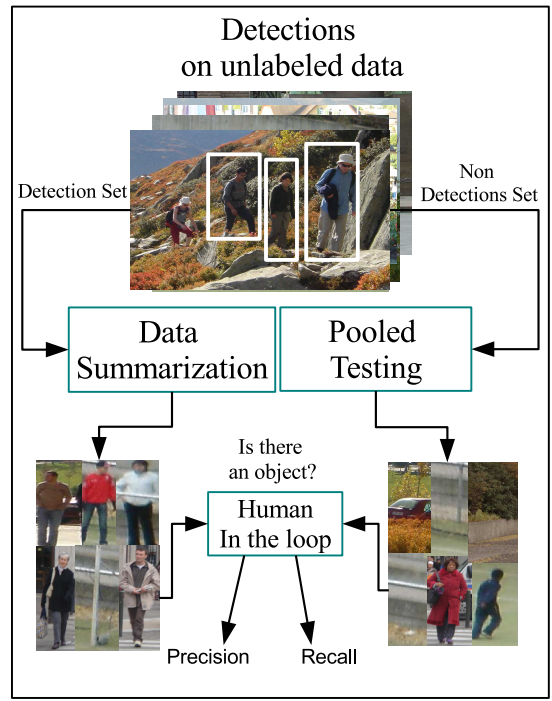}
\caption{\small{Our system estimates the performance of an object detector on unlabeled data using feedback from a human in the loop.}}
\label{fig:overview}
\vspace{-0.2in}
\end{figure}
Object detection is one of the most popular and practical applications of computer vision today. Even though it has been extensively studied for over two decades, it continues to be a challenging problem. The difficulty in solving the problem is attributed to several factors such as change in lighting, pose, viewpoint, size, shape and texture. However, there have been significant advances in building robust object detectors that are finally making them commercially viable in applications for home security, surveillance, online social networks, online shopping etc. In the near future, it is conceivable that even individual end-users will wish to purchase specific detectors for their specific use. For example VMX \cite{vmx_www}, is being developed as a ``computer vision as a service" that allows one to outsource detectors and classifiers while focusing on their applications. In such a scenario, how does one allow a user to get a quick sense of how well the detector generalizes to their specific requirement, as opposed to interpreting the ROC curves of detectors on certain test-beds when the test-beds themselves may or may not be reflective of a user's intended deployment scenario. Further, there has been an explosion of the number of detectors in the past decade promoted by challenges such as PASCAL VOC \cite{Everingham10} where in the past three years alone, $43$ methods completed the challenge ($2010-22, 2011-13, 2012-8$) \cite{VOCwww}. Traditionally, rigorous evaluation of detectors requires extensive ground-truthing, which by its very nature restricts an end-user from judging how well the detector will generalize outside the test-bed. Theoretical bounds offered by statistical learning theory \cite{vapnikBook1998} offer guidance to algorithm developers, however the bounds themselves are generally quite weak and provide little useful information to end-users. The question we seek to answer in this paper is, how does one get a quick sense of a detector's performance for a specific application without the need for extensive groundtruthing?

In this paper, we propose an interactive system by approaching this as a problem of smart sampling to estimate object/non-object proportions, a block diagram of the system is shown in fig \ref{fig:overview}. Let $U$ be the set of all possible image patches among user provided images. Denote by $D\subset U$, a`detection-set' which were reported as containing the object of interest by the detector and $\overline{D}$, the complement of the detection set. The challenge is to then sample informatively from $D$ and $\overline{D}$ in a way that quickly allows us to estimate class proportions. Proportions are useful because they are independent of the size of dataset, and two of the important measures in detection theory, precision and recall are expressed as fractions. So given a large collection of unlabeled images, which are the best samples to pick that preserve ``class proportions''? The answer is not trivial, random sampling methods are accurate only when class proportions are nearly equal. Assuming this step is solved, the next step is to ground truth the chosen samples by asking a simple yes/no question to a human, based on which we can estimate the proportion of objects within the set of detections. Performance for a detector is always measured relative to ground truth, therefore in the absence of labels we seek to estimate ``perceived performance'' and show that this is indeed close to actual performance measured with ground truth when using the proposed sampling procedures. We adopt and extend data sampling strategies and present a coherent framework to estimate false positives (incorrect detection) and a probabilistic estimation approach known as group testing to determine the false negatives (missed detection). We treat the detector as a black box and operate directly on the detections.

The main idea is the following, first  for different detection thresholds ($\gamma$) we feed the set of detections to a data summarization algorithm that identifies a small number of samples based on preserving some notion of information. These samples are generated for each $\gamma$  and presented to the human who gives them a 1/0 label for object/non-object. If the samples are chosen well, the class proportions are expected to be preserved giving us an accurate estimate of precision (portion of total detections that were true).

 Recall is the other important measure, which is the portion of total objects that were detected. Once we estimate precision, we only need to estimate the number of false negatives from $\overline{D}$ to obtain recall.  Unfortunately obtaining the complimentary set $\overline{D}$ is not easy and it is further complicated by the fact that the objects are very similar to background in the feature space. This results in the failure of any feature based approaches such as summarization and clustering. To address this issue we use the theory of statistical group testing that is employed to study estimation of disease propagation among large populations of plants, animals or humans. Essentially, we pool in multiple images before `testing' them with the human for the presence of an object. This information is used to obtain estimates on the number of false negatives within $\overline{D}$.

{\bf Contributions} 1) We present the first system that allows a user to estimate detector performance without the need for extensive groundtruthing. 2) We achieve this by employing probabilistic techniques such as pooled testing and smart sampling methods such as summarization by keeping a human in the loop. This will allow users to ``test-drive'' object detectors in the future while minimizing time and effort required for labeling. 3) Our system also gives the best performance estimate compared to simpler baselines with the least amount of human effort.

\section{Relevant Work}
In this section we outline the important work that has been done in the different fields of research relevant to our system. 

{\bf Object detection} is the problem of putting a bounding box around an object within an image. Generally speaking, it involves running a window over the image at different scales to check for the highest match with some model representation of the object obtained from training data. In this work we look at pedestrian detection with three detectors, namely integral channel features (ChnFtrs) \cite{DollarTPB09}, Multiple features with color self similarity (MultiFtrs+CSS) \cite{WalkMSS10} and aggregate channel features (ACF)\cite{DollarBP10}, all of which have performed well on many standard datasets. For a detailed comparison of different pedestrian detectors on several datasets we refer the reader to \cite{DollarWSP12}. There are several variations of each detector mainly varying in the choice of local descriptors or using extra information such as color, context etc. A popular feature used in most detectors is Histogram of Oriented Gradients (HOG) feature introduced by the Dalal-Triggs person detector \cite{Dalal2005}. Our system operates on the detections so it is applicable to any kind of object detector.

{\it Determining true positives with labels:} Once a bounding box has been predicted, an overlap measure determines if the detection is a true or false positive based on its overlap area with the ground truth. Like any detection problem, the main quantities that determine the quality of a detector are the number of false positives and false negatives. An ideal detector would have both quantities to be low over various detection thresholds.

{\bf Video summarization} attempts to identify the most concise representation of a video while maximizing `information' content. Usually, algorithms for summarization differ in their definition of information keeping the main application in mind. Recent methods in summarization have employed high-level contextual features and user input to present summaries\cite{LeeGG12}. The summarization algorithm we seek for this system is a more general form of clustering and must be able to deal with image based features rather than video based ones such as flow, object tracks etc. Precis \cite{ShroffTC10,ShroffTC11} is one such algorithm that can work with image based features. Precis looks at maximizing `diversity' and minimizing representational error among the data points. 

{\bf Human in the loop systems } Many difficult problems in vision including segmentation, summarization, and even learning have been shown to benefit by human-in-the-loop systems. The motivation for such systems is to actively minimize human effort in tasks that are otherwise laborious such as image or video annotations, active video segmentaion etc \cite{VijayanarasimhanG09}. Typically a human is asked a set of questions, the answers to which are used as feedback to the system to reduce the search space. For example, Branson et al. \cite{BransonWSBWPB10} develop a system that performs multi-object classification and uses pre-processing to identify a small set of important questions to ask the human, which then enables learning. This allows the authors to perform classification of subtly different classes which would have been very hard otherwise. Since we are working with unlabeled data, the role of the human is critical in determining the final estimated performance. Humans also have an extraordinary ability to identify objects using multiple cues \cite{Spelke1990}, which can be taken advantage of for image pooling.  

{\bf Group testing} Pooled or group testing has it origins in the area of groundtruthing agricultural/biological samples, and its mathematical formulation has since been applied more broadly. The underlying premise is that individuals (blood samples, seed kernels etc.) are first pooled together into groups and the groups are tested for the presence of infection. When infection rate is low, there is an overwhelming evidence that pool testing can confer substantial benefits in terms of testing/labeling effort when compared to testing individually \cite{Hepworth2013,Chen1990}. A complete discussion of the assumptions, robustness of the estimates as a function of group size, errors in testing etc. can be found in \cite{Chen1990}. These ideas have been applied to estimating disease propagation in large populations of plants, animals or humans \cite{Hepworth2013}. 

{\bf Image retrieval from large datasets} The techniques described here are very similar to methods that search for similar images within large data collections. For example hashing methods such geometric min-hash \cite{ChumPM09}, and tree based methods such as \cite{NisterS06} could potentially be used to find the number of true detections in $D$, but this requires one to have trained examples to learn the underlying data structure and we set out to work with unlabeled data with no prior training. It doesn't help much for the negative set either since the feature descriptors are expected to be poor due to occlusions, shadows etc., which caused the detector to miss them the first time.

\section{Approach \& Implementation}
 In this section we will describe our system in detail and the theory behind estimating detector performance. The system presumes that a detector is used off-the-shelf, and is completely agnostic to the detector's algorithm or features. A note on terminology used here, the term `class' has no meaning in unlabeled data, but we refer to it as the label that would have been assigned by a human. Therefore, our estimate of quantities are `perceived' but as we show, they are close to the true underlying values.

{\bf Measuring Performance:} Knowledge of the number of false positives (FP) and false negatives (FN) provides sufficient information in most cases to judge the performance of a detector. Some applications may require the false positives per image (FPPI) to be restricted below a specification before implementing it into their system, but our current system does not allow us to estimate FPPI, so we will focus on the two main quantities FP, FN. 

\subsection{Sampling from the detections set to estimate perceived precision}
Interactively estimating the proportions of different classes (objects, non-objects here) from a large set of images essentially boils down to smart sampling, i.e. the problem would be solved if we were to pick the right proportion of samples from each class.  Picking random samples is of little help in cases with skewed class proportions i.e., when the detector is operating at high precision. Given the fact that any detector based application is commercially viable only when the detectors are mature enough to have high precision and recall, the random sampling method is not suitable. One could use clustering algorithms such as K-means (K-medoids) and use the cluster centers as the samples, but this suffers from the same pitfalls as random sampling and does only marginally better. 
\begin{figure}[!htb]
\includegraphics[trim = 0 15mm 0mm 0, clip, width=8.5cm]{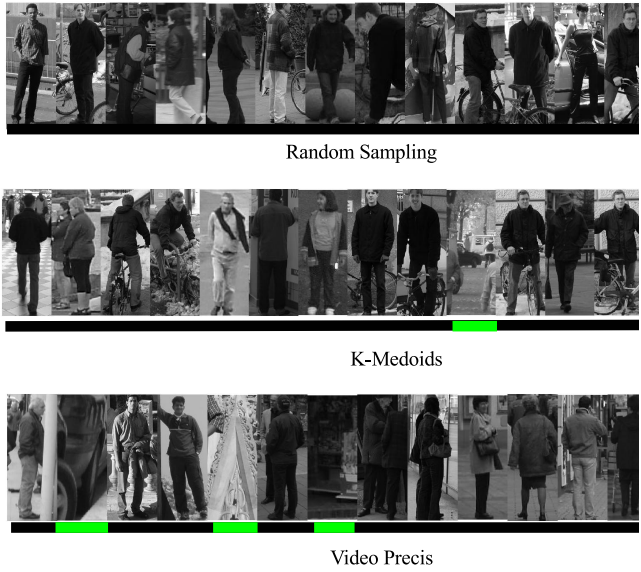}
\caption{\small{Random sampling and K-medoids overestimate the proportion of pedestrians  because they fail to retrieve samples from the smaller class of non-objects. Precis'\cite{ShroffTC11} samples give a much closer estimate to the ground truth. Here the non-objects are highlighted for a fixed $\gamma$ from the INRIA pedestrian dataset \cite{Dalal2005}, the perceived precision for random, K-means, Precis are 1, 0.9375, 0.8125 respectively compared to the actual precision = 0.7.}}
\label{fig:precisBest}
\vspace{-0.1in}
\end{figure}

Let the set of detections at a specific threshold $\gamma$, be denoted as $D$, define $D_{fp}, D_{tp} \subset D$ as the set of false and true detections respectively and $D_{fn} \subset \overline{D}$ as the set of missed detections or false negatives. Then $ \left\vert{D_{fp}}\right\vert<<\left\vert{D_{tp}}\right\vert$ in most cases, where $ \left\vert{.}\right\vert$ denotes the cardinality of a set. How does one sample informatively from $D_{tp}~\&~D_{fp}$ when they are not explicitly known? The first step is to extract a feature vector from each image patch, we use the histogram of oriented gradients for this purpose. The problem now is to identify the $K_\gamma(<<\left\vert D\right\vert)$ best samples, denoted by $S$ from the data feature matrix $X$ such that the proportion of elements in $D_{tp}$ and $D_{fp}$ are preserved. Since the number of false positives is far smaller than the number of true positives using algorithms such as K-medoids will result in picking centers that are mostly from $D_{tp}$ since it only looks to optimize representational error, resulting in poor proportion estimates. Instead we use Precis \cite{ShroffTC10,ShroffTC11} to pick samples well distributed across each set. Precis solves for a set of samples $S = \{F_{i}\vert i = 1,2\dots K_\gamma\}$, that minimize representational error (`coverage') denoted by $rep(S,X)$ and maximize diversity between chosen samples, denoted by $div(S,X)$. These quantities are given by 
\begin{align}
rep(S,X)& = tr\left[\sum_i \sum_{F_k\in V_{i}}(F_k - F_{i})(F_k - F_{i})^T\right],\\
 div(S,X)& = tr\left[\sigma_i(F_{i}-\overline{F})(F_{i}-\overline{F})^T\right],
\end{align} 
where $V_{i}$ is the partition corresponding to $F_{i}$, i.e. $V_{i}$ are the elements of $X$ that are closer to $F_{i}$ than any other element in $S$. $\overline{F} = \frac{1}{K_\gamma}\sum_j F_{j}$ is the mean of $S$. This results much better sampling across $D_{tp} ~\&~ D_{fp}$ as shown in fig \ref{fig:precisBest}.

\paragraph{Homogenizing the detection set} In general, samples chosen from $D_{fp}$ are not well behaved in the feature space because they contain everything but the object of interest. This can throw off any metric based algorithm such as K-medoids due to outliers and Precis due to increased diversity etc., resulting in poor samples. To ensure better sample selection, we transform the data to a space where $D_{fp}$ and $D_{tp}$ are well separated. This transform, denoted by $\mathcal{T}$, needs to be estimated only once per object class and is obtained by minimizing a cost function using data points from $O_T$, a set of image patches with objects and $O_F$ a set of image patches without objects. Since both of these sets are unknown for unlabeled data, we will approximate them by picking $O_T = D_{\gamma_H}$ which is the detection set at a high confidence threshold $\gamma_H$ and $O_F = \overline{D}_{\gamma_L}$ the non-detection set at a low confidence threshold $\gamma_L$. This is a reasonable approximation since at $\gamma_H$ one expects very few false positives and at $\gamma_L$ one expects very few false negatives.

Typically, we want to ensure that $d_{\mathcal{T}}(x_i,x_j)\leq u$, for $(x_i,x_j) \in O_T$ and $d_{\mathcal{T}}(x_i,x_j)\geq \ell$ for $x_i \in O_T ~\&~ x_j \in O_F$ for some $u,\ell>0$. The loss function(s) depend mainly on the Euclidean norm on the transformed space, $X^T\mathcal{T}^T\mathcal{T}X$. An example of such constraints  are 
\begin{align}
c_1& = \mbox{max}(0,d_\mathcal{T}(x_i,x_j)-u) \mbox{ for } (x_i,x_j) \in O_T, \\
c_2& = \mbox{max}(0,\ell-d_\mathcal{T}(x_i,x_j)) \mbox{ for } x_i \in O_T ~\&~ x_j \in O_F.
\end{align}

Other loss functions can be used depending on the application. A constraint on $\mathcal{T}$ is imposed by a regularizing condition $reg(\mathcal{T})$ (such as $\mathcal{T}\geq {\bf 0}$). A general cost function for $m$ loss functions is given as follows:

\begin{equation}
J_1(\mathcal{T}) = reg(\mathcal{T})+ \lambda \sum_i^m c_i.
\label{eqn:metric}
\end{equation}
This is very similar to the metric learning problem and other examples of loss functions and regularizers can be found in \cite{Kulis2013}.
In this paper, we use Information Theoretic Metric Learning (ITML) \cite{DavisKJSD07} to find the optimal $\mathcal{T}$. The estimated transform is independent of the particular samples and is estimated once per object class.
Once we have the optimal $\mathcal{T}$, we solve for the best set of samples, $S$ using Precis which minimizes the cost function:
\begin{equation}
 J_2(S) = \alpha~rep(S,\mathcal{T}X) +(1-\alpha)~div(S,\mathcal{T}X).
\end{equation}

Finally, the set of samples are shown to a human who marks false positives. Typically $K_\gamma$ is chosen to be about $5-10\%$ of the detections at $\gamma$. From this, estimated precision is calculated as  :  $\widehat{P}_\gamma = 1 -\frac{\widehat{FP}_\gamma}{K_\gamma}$ where $\widehat{FP}_\gamma$ is the perceived number of false positives at $\gamma$.

\subsection{Pooled testing of samples from the detection complement}
Metric based approaches such as clustering, Precis and random sampling severely underestimate the number of false negatives due multiple reasons: 1)The proportions of objects to non-objects are much smaller ($\sim 1-5\%$ as compared to at least ~10\% for false positives) and 2) the fact that the detector missed them, indicate that they are very hard to distinguish from the background in the feature space. For this reason, we use probabilistic methods such as pooled testing that allows us to quickly estimate the number of false negatives within the image population. Pooled testing is used often to estimate disease propagation within large populations of plants and animals. It is especially useful during the outbreak of an epidemic where there is a need to quickly estimate the extent of damage before taking action.

 In the estimation problem, the primary goal is to estimate $p$, the individual probability of infection by testing pools of size $s$, of individual samples for the infection. There are different ways of testing pools, we use the inverse binomial sampling method \cite{Pritchard2011} which provides the fastest estimates. In this method, suppose that $T$ pools need to be tested before $n$ positive samples have been identified, then the maximum likelihood estimate of $p$ is given by 
\begin{equation}
\hat{p} = 1 - \left(1-\frac{n}{T}\right)^\frac{1}{s}.
\label{eqn:est_p}
\end{equation}

Better estimates of $p$ that take into account unequal pool size, inaccurate testing etc. can be found in \cite{Chen1990,Hepworth2013}. 

Before estimating the number of false negatives using \eqref{eqn:est_p}, we need to accurately obtain $\overline{D}$, where the image patches are such that each one contains at most a single object, this is not easy since objects can be of different scales within the same image and there may be multiple objects with occlusions. Assuming we have an accurate $\overline{D}$, our problem is to estimate the number of patches with objects in them. This is a catch-22 situation since it is the detection problem all over again, therefore we resort to probabilistic methods like pooled testing. In order to obtain $\overline{D}$, we make the assumptions that most objects in an image are of approximately the same size and that the objects are not occluding each other. This allows us to use bounding boxes obtained from the detector as an indicator of scale for an object within the image. Next the bounding box is used to divide the image into smaller patches, excluding regions previously detected. This generates $\overline{D}$ which contains a large set of image patches with a small number of objects (usually $\sim1-5\%$). In cases of multiple detections, we choose the box with the average height and width. 

\begin{figure}[!htb]
\includegraphics[trim = 25mm 30mm 10mm 10mm, clip, width=8.5cm]{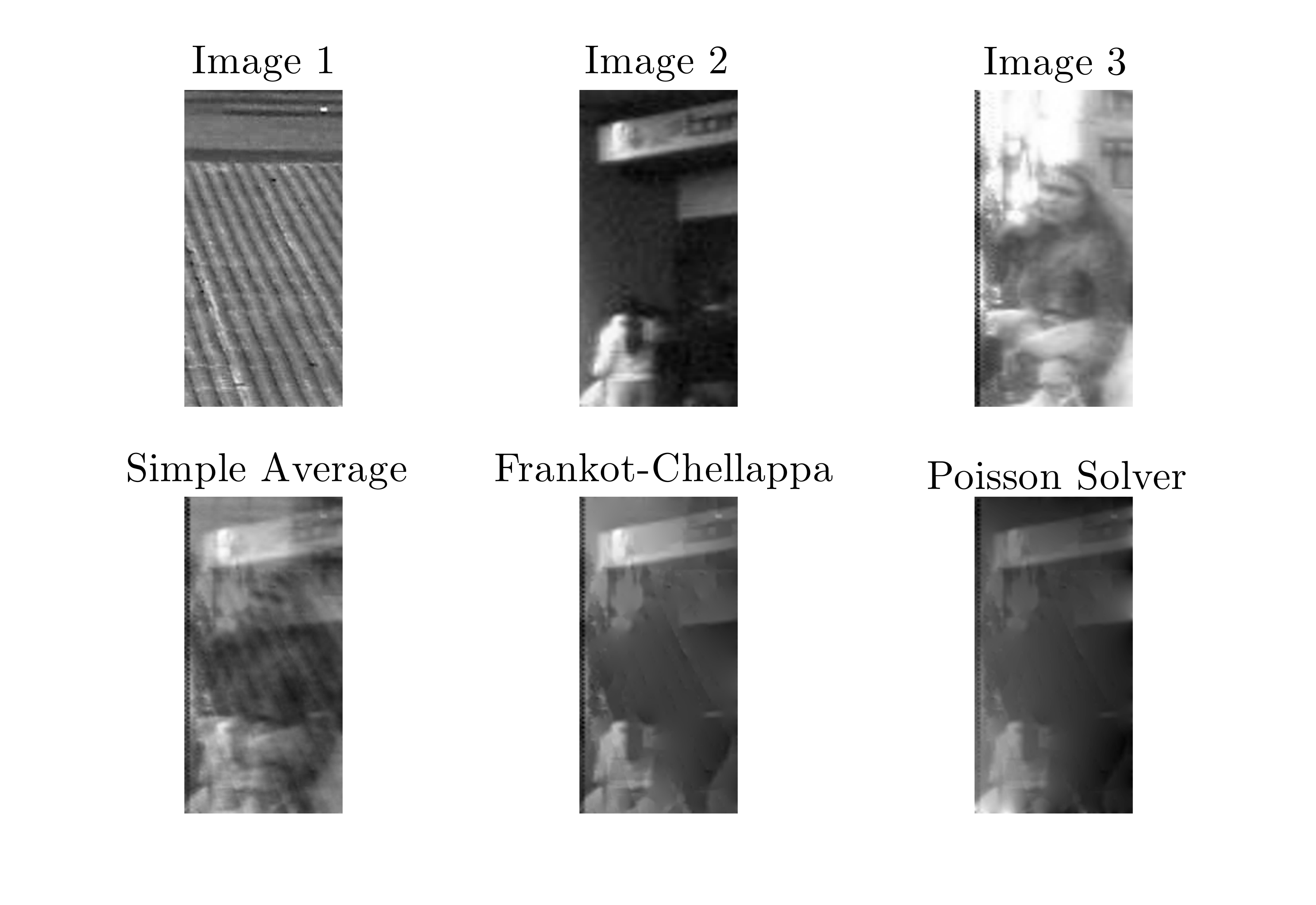}
\caption{\small{{\bf Image Pooling -} This example shows a simple averaging scheme performing perceptually better than gradient domain methods, the original images are shown in the first row. Although gradient domain methods are attractive for image fusion, they may produce artifacts if they are not tuned correctly and work poorly on low-res images.}}
\label{fig:imfusion}
\vspace{-0.2in}
\end{figure}

\paragraph{Approaches for Image Pooling}
The images need to be pooled together in a way that their individual characteristics (such as edges) are preserved. There are multiple ways to achieve this, and can be chosen specific to the application. In applying this to pedestrian detection we looked at three techniques namely, a) spatial averaging of two images  b) Gradient domain averaging using the Frankot-Chellappa algorithm \cite{FrankotChellappa1988} and c) Poisson image blending \cite{Perez2003}. Gradient domain techniques are appealing to use in this context since they can be tuned to preserve high frequency content (such as edges, corners) and leave out background. Concatenation of multiple images can also be effective instead of pooling them, but one must consider the trade-off between the time it takes for humans to analyze a single pooled image versus multiple individual images, currently there is no research suggesting either way and a thorough evaluation of this is left for future work. A qualitative comparison of the these methods is shown in fig \ref{fig:imfusion}. Based on our experience with multiple trials, we found that the simple spatial average pooling was suffcient for seeking interative feedback from a user. Thus, we report results via the simple spatial average pooling in experiments. 

\begin{figure*}[!htb]

  \subfigure[]{\includegraphics[clip = true,trim=0mm 0mm 0mm 0mm,width = 5.8cm,height = 5cm]{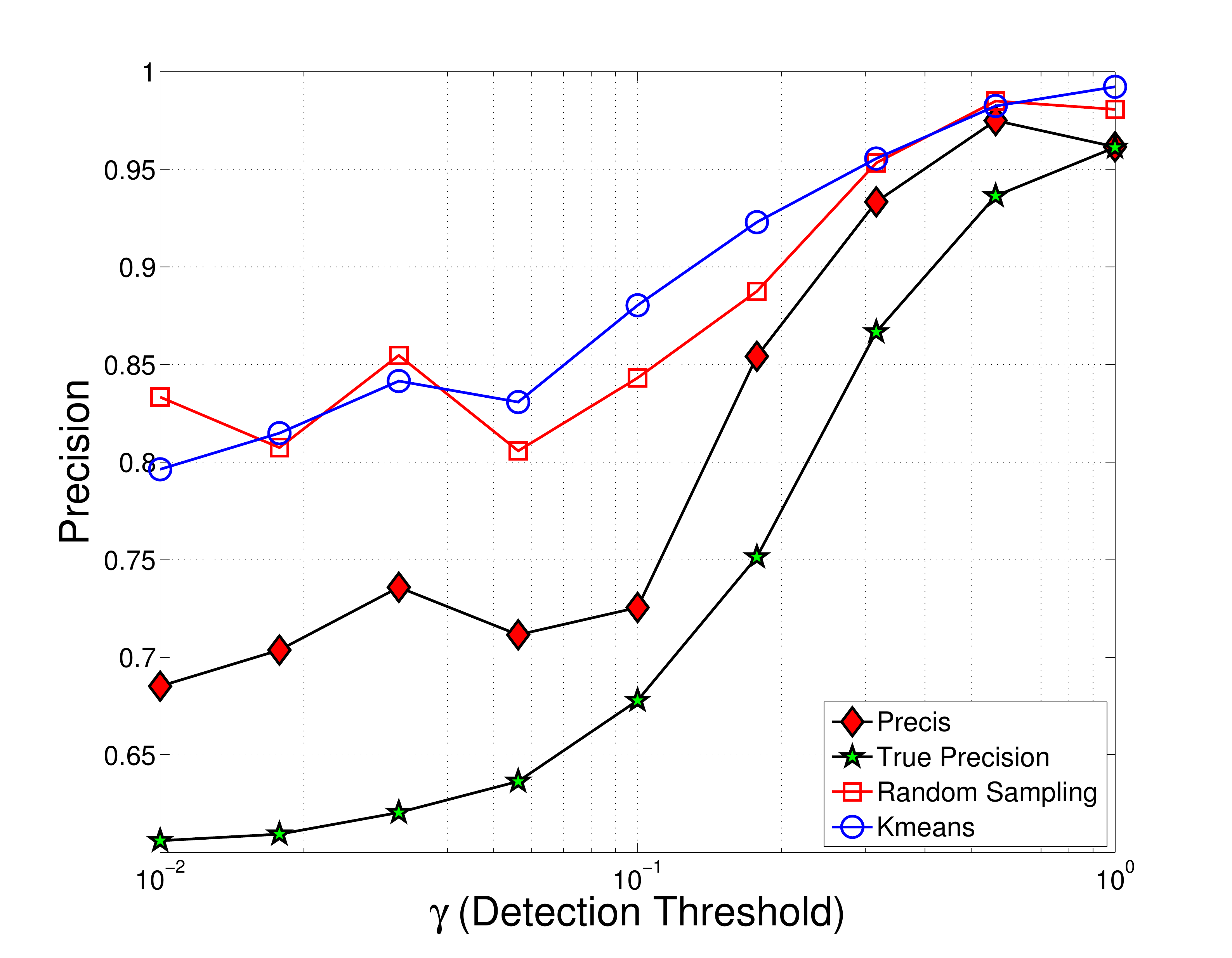}
 \label{fig:ACF}
 }
\subfigure[]{  \includegraphics[clip = true,trim=10mm 0mm 0mm 0mm,width = 5.8cm,height = 5cm]{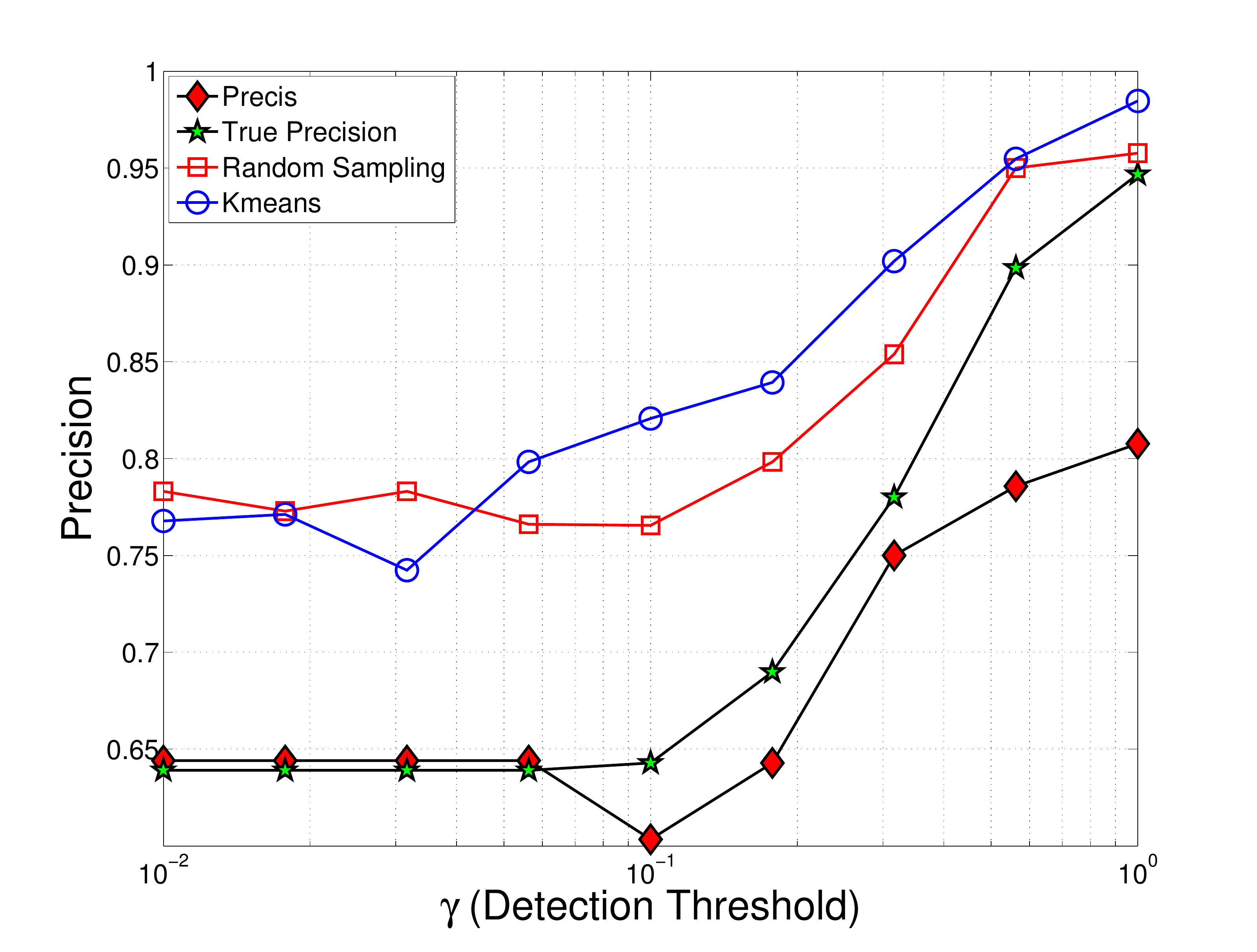}
    \label{fig:ChnFtrs}
 }
\subfigure[]{ \includegraphics[clip = true,trim=10mm 0mm 0mm 0mm,width = 5.8cm,height = 5cm]{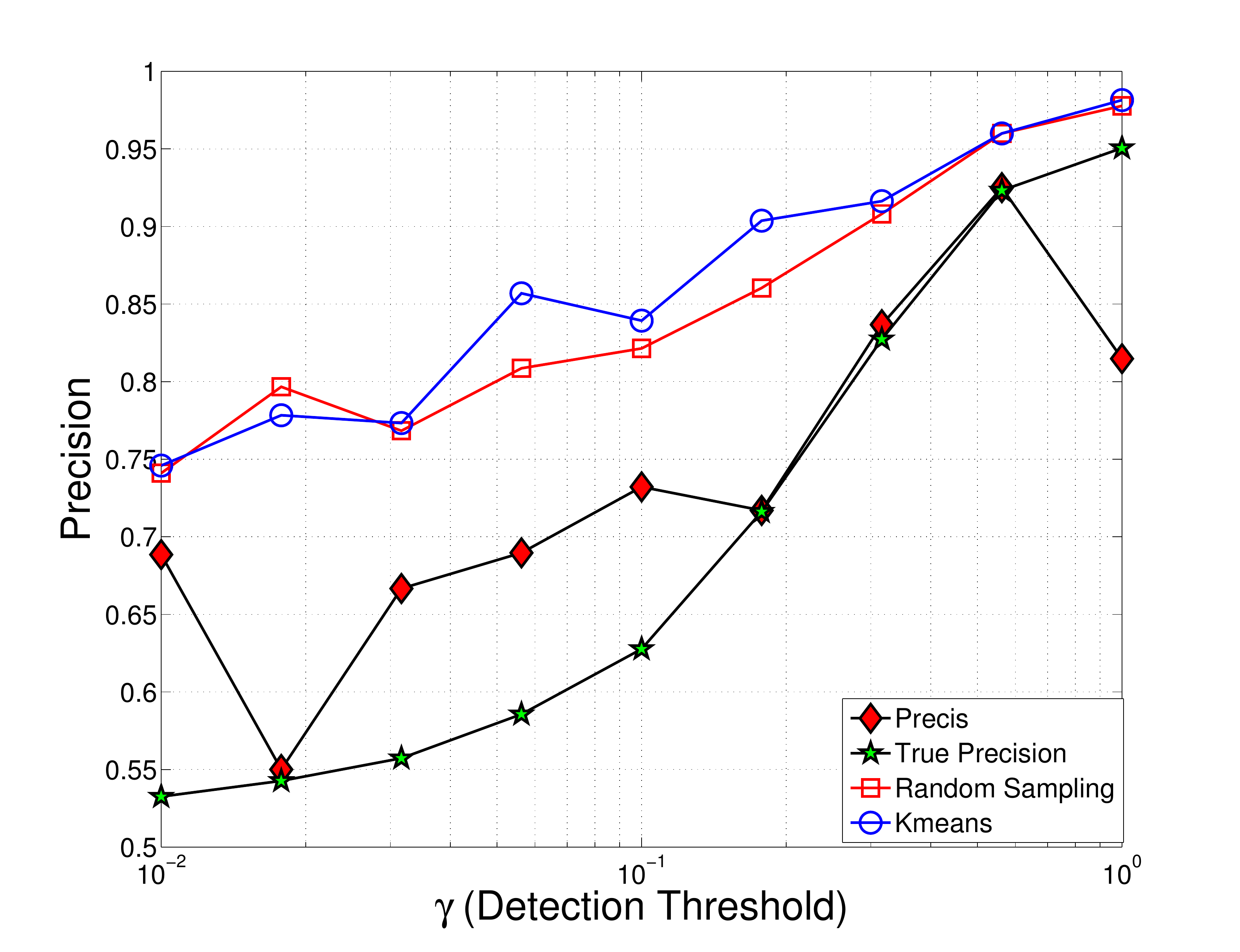}
\label{fig:Multi}}

  \subfigure[ACF Detector]{\includegraphics[clip = true,trim=0mm 0mm 0mm 0mm,width = 5.8cm,height = 5cm]{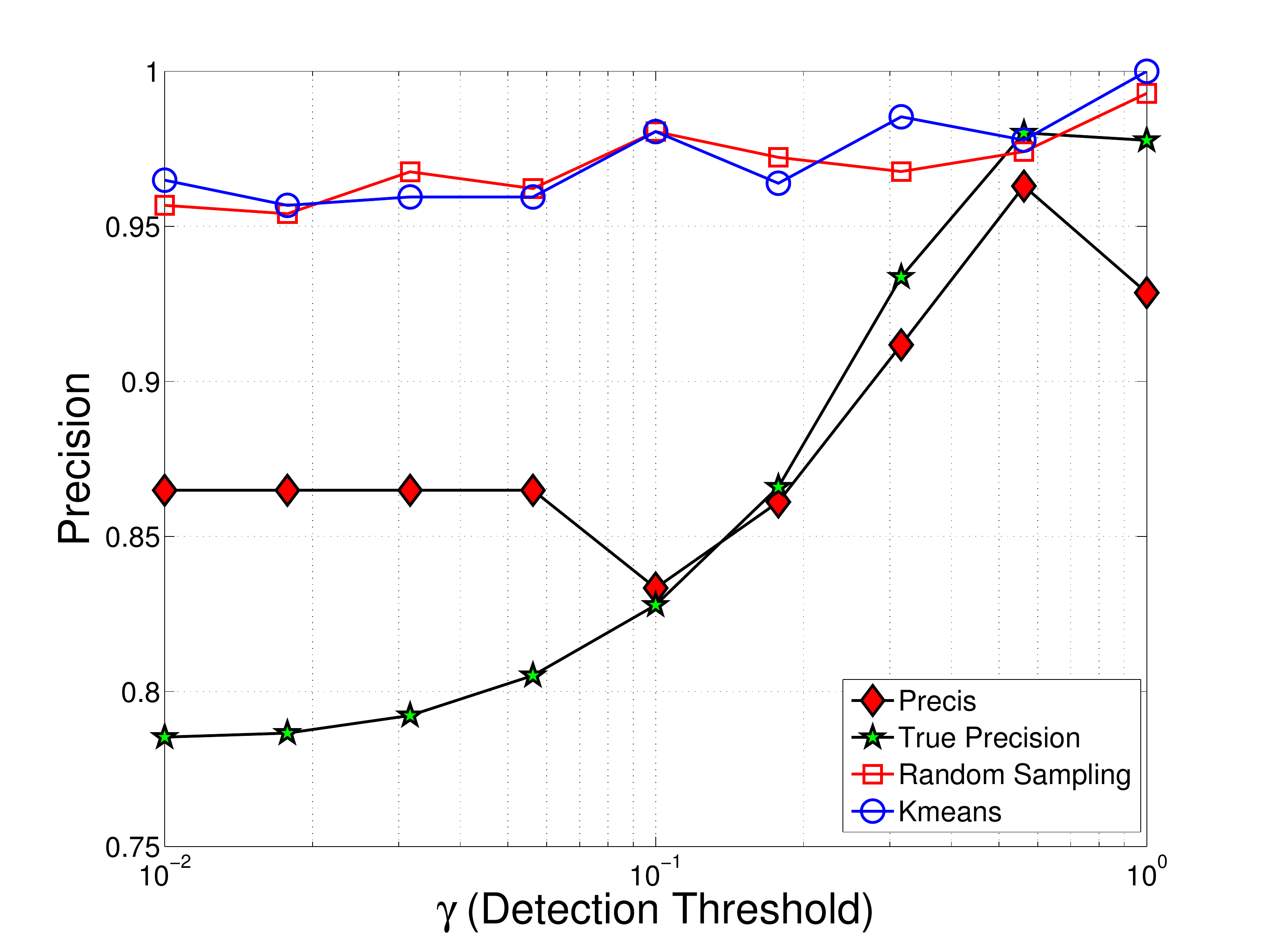}
    \label{fig:ACF2}
   }
\subfigure[ChnFtrs Detector]{\includegraphics[clip = true,trim=10mm 0mm 0mm 0mm,width = 5.8cm,height = 5cm]{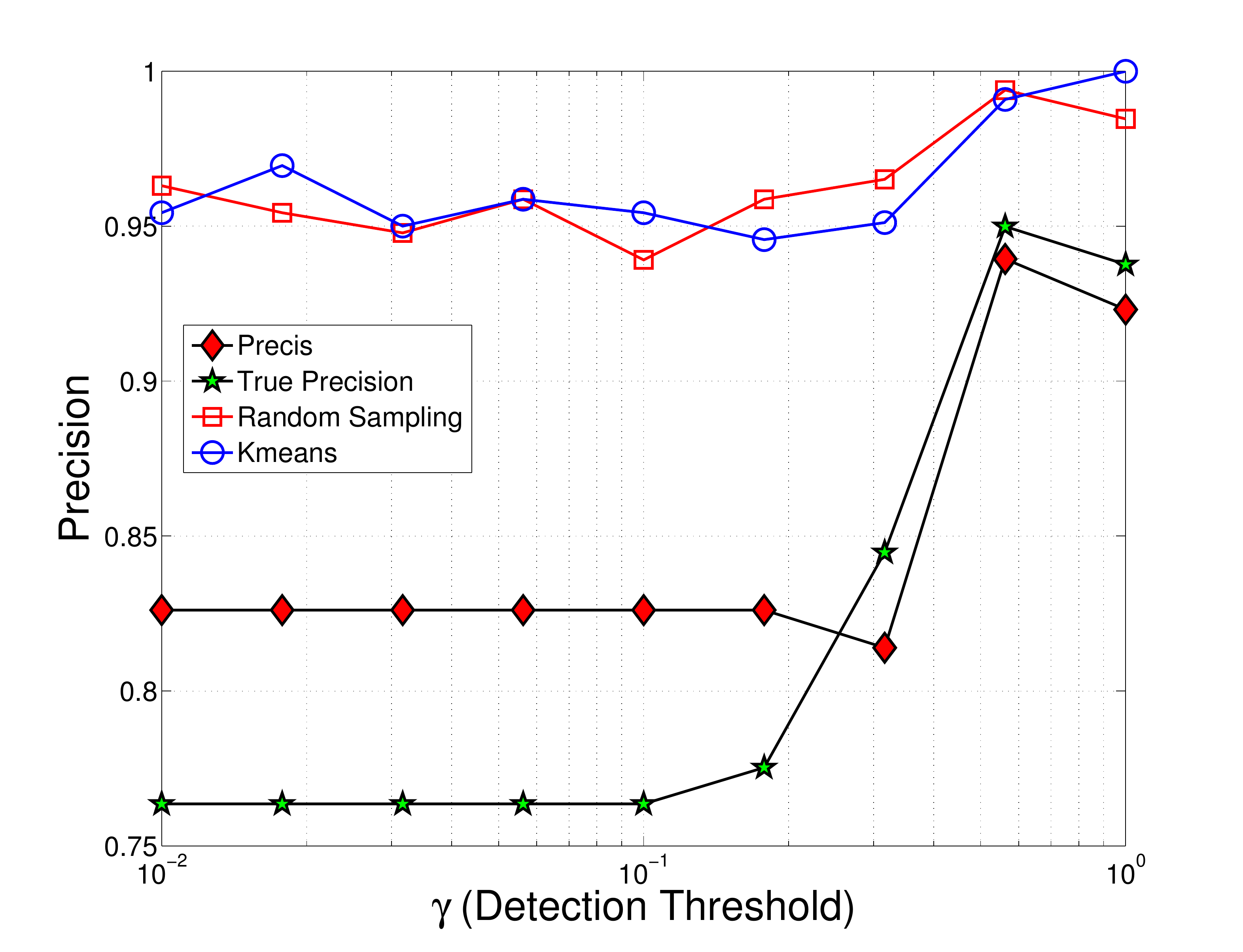}
    \label{fig:ChnFtrs2}
 }
\subfigure[MultiFtr+CSS Detector]{ \includegraphics[clip = true,trim=10mm 0mm 0mm 0mm,width = 5.8cm,height = 5cm]{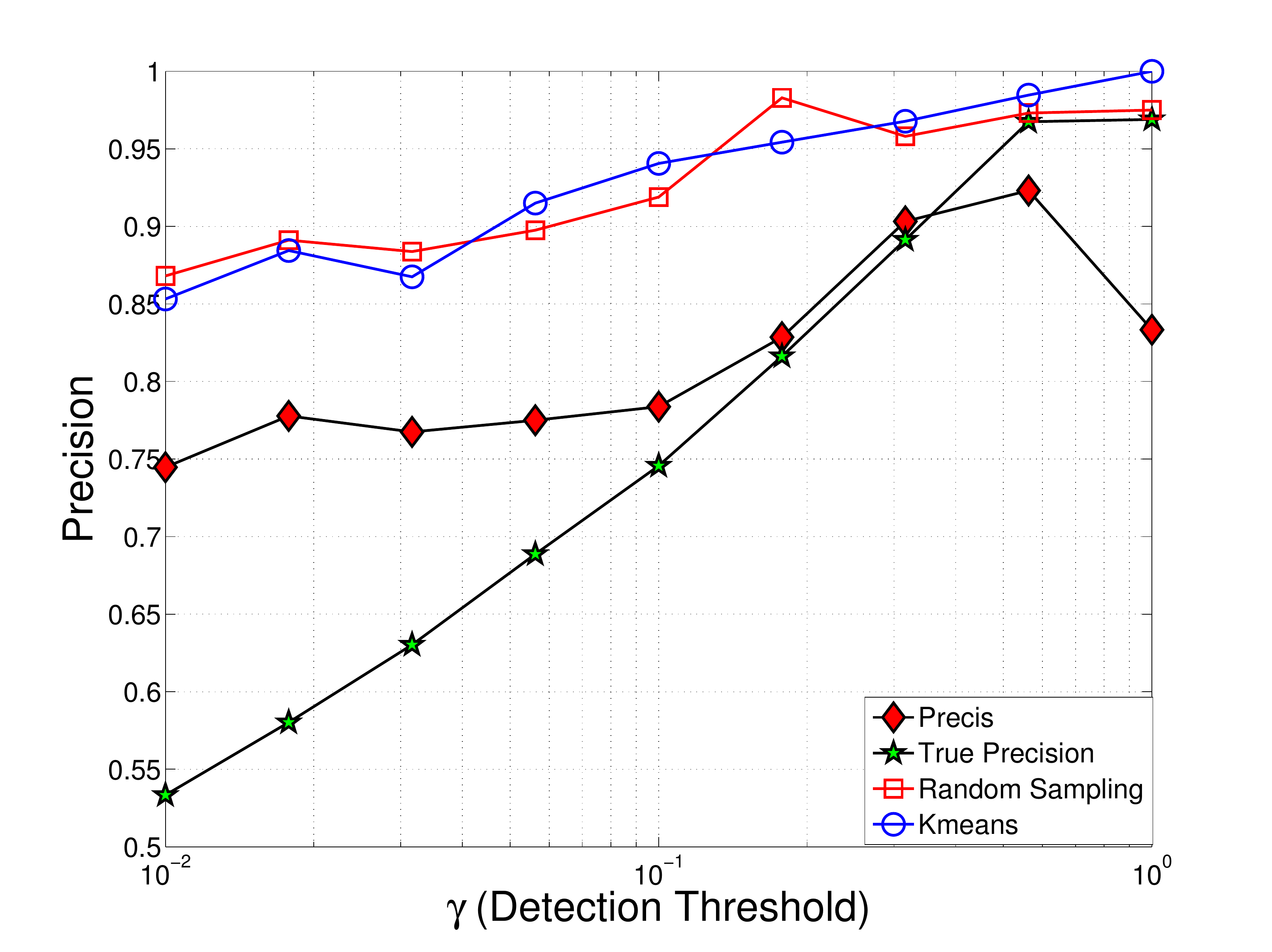}
\label{fig:Multi2}
}

\caption{\small{Comparing the precision of ACF Detector \cite{DollarBP10}, the ChnFtrs Detector \cite{DollarTPB09} and MultiFtr+CSS Detector \cite{WalkMSS10} on the INRIA Pedestrian Dataset \cite{Dalal2005} (\ref{fig:ACF},\ref{fig:ChnFtrs},\ref{fig:Multi}) and ETH Dataset \cite{eth_biwi_2008} (\ref{fig:ACF2}, \ref{fig:ChnFtrs2}, \ref{fig:Multi2})  using just 10\% of the total number of detections at each $\gamma$. The values for K-means and random sampling are averaged over 10 iterations as compared to a single iteration of Precis.}}
	\label{fig:est_precision}
\vspace{-0.2in}
\end{figure*} 

\section{Experiments}
In this section we demonstrate our system and its performance estimates compared to performance obtained using ground truth. We show results on the INRIA Pedestrian dataset \cite{Dalal2005} and the ETH Pedestrian Dataset \cite{eth_biwi_2008}, which are popularly used for pedestrian detection. The INRIA dataset is relatively simple with 2-4 pedestrians on average per image and of roughly the same scale. ETH dataset is more complex, with 10-12 pedestrians per frame and of varying sizes, it is acquired using a moving camera on a crowded street. We estimate the performance of three detectors namely Aggregate Channel Features (ACF)\cite{DollarBP10}, Integral Channel Features (ChnFtrs)\cite{DollarTPB09}, MultiFtr+CSS \cite{WalkMSS10} on these datasets. 

\subsection{Interactively test-driving predestrian detectors}
There has been great success in the field of pedestrian detection with a large number of excellent detectors proposed in the past few years. A good evaluation of the state of the art methods is available in \cite{DollarWSP12}, the authors have also made available the detection outputs for several detectors and datasets on the web, we choose 3 detectors and 2 datasets to demonstrate the proof of concept.  

{\bf Note on using labels to generate results} Samples obtained from a given sampling algorithm (random, K-medoids or homogenized Precis) are shown to a human to identify the number of false positives, using which the `perceived' precision rate is calculated as explained earlier. Since we work with labeled datasets, we simulate the process with the given label which enabled much faster results for the sake of testing and proof of concept. However multiple duplicate detections around an object are considered false positives when evaluating with ground truth where only the detection with the highest overlap is chosen as the true detection and the rest are considered false. It is impossible to tell duplicates apart by looking at the detections from unlabeled data.  To account for this during run time, we consider all the duplicate detections to be true detections even though they are actually false detections. 

We used the detections for two datasets and estimated precision for different $\gamma$, the results for estimates based on only $10\%$ of the total number of detections are shown in fig \ref{fig:est_precision}. It is seen that the proposed homogenized Precis outperforms K-means and random sampling significantly. However, in most cases there is still a positive bias in the estimate which we attribute to the duplicate false positives that are impossible to distinguish without true labels. 

The K-means and random sampling estimates are averages over $10$ iterations, because they tend to be noisy, meaning this would require the human to look at 10 times the number of images used by Precis before getting a comparable estimate. In picking $K_\gamma$, smaller is lesser work for the human, and we also found that there is not a significant difference in performance in most cases. 

\begin{figure}[!htb]
\includegraphics[trim = 10 0mm 10mm 10, clip, width=8.7cm]{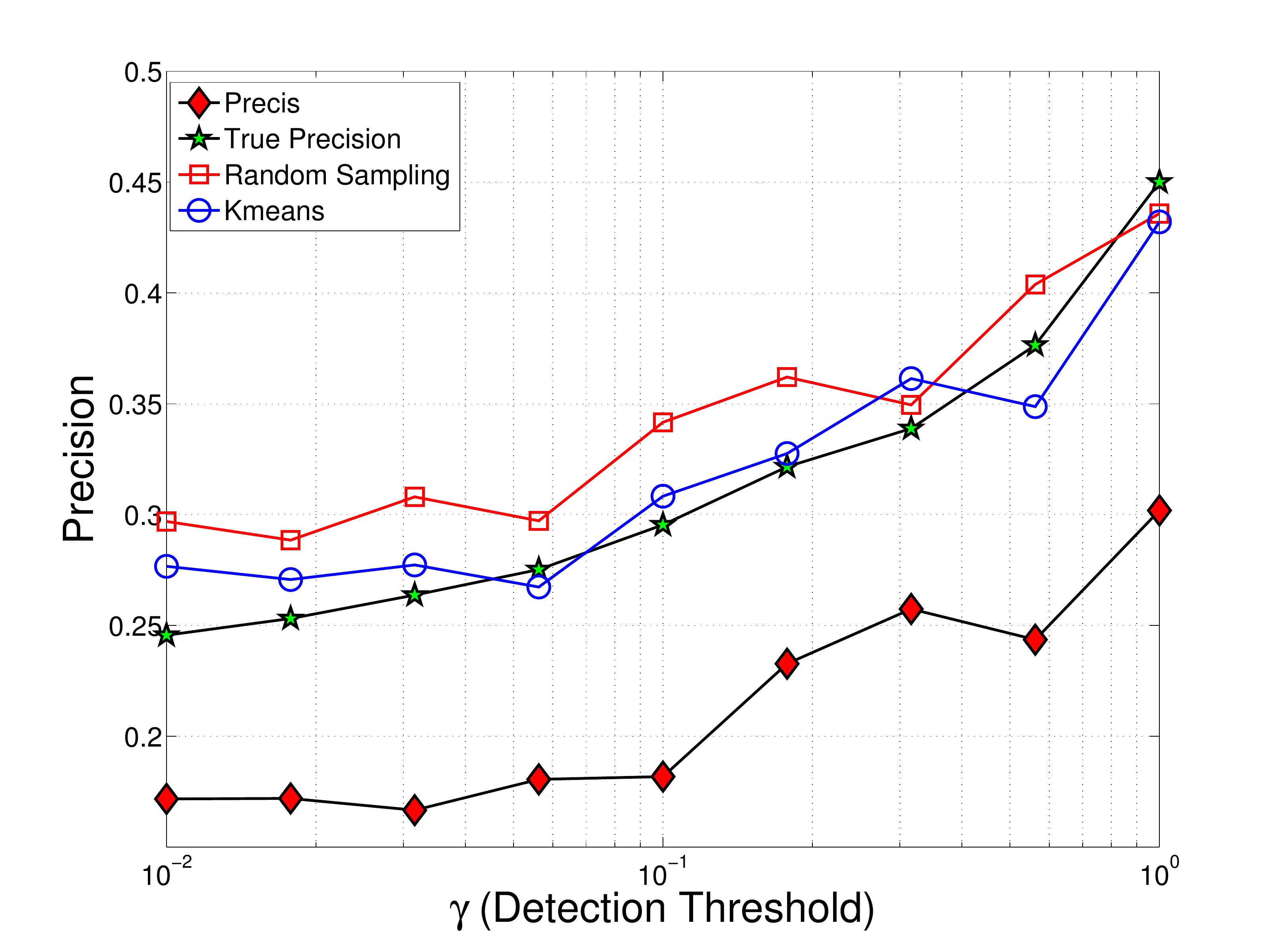}
\caption{\small{Precis fails to estimate precision accurately for detectors that perform poorly, but this is hardly an issue since poor detectors are never preferred in applications. The precision estimates are shown for the HOG+SVM detector \cite{Dalal2005} on the INRIA dataset.}}
\label{fig:hogFail}
\vspace{-0.2in}
\end{figure}

\subsection{Failure cases}
The performance estimate using the Dalal and Triggs detector \cite{Dalal2005} is shown in fig \ref{fig:hogFail}. Precis has a poor estimate in this case compared to random sampling and K-means. This is because the general performance of the detector is far lower than state of the art detectors today, which results in many more false positives. Precis by construction is sensitive to diversity and often tends to pick many more false positives than pedestrians resulting in under estimating precision severely. Fortunately, this condition is not limiting since a detector is not used commercially until it is high performing. 

\subsection{Estimating the number of false negatives}
Recall is the portion of objects that were correctly detected and can be completely described by $\gamma,P_\gamma,N_\gamma,\vert D_{fn}\vert$ where $\gamma$ is the detection threshold, $P_\gamma$ is the precision, $N_\gamma$ is the number of detections and $\vert D_{fn}\vert$ is the number of false negatives or missed detections. All the quantities are known except $\vert D_{fn}\vert$ so we will focus on estimating the number of false negatives, which gives us all the information to calculate Recall.

\begin{figure*}[!htb]
  \subfigure[]{\includegraphics[clip = true,trim=0mm 0mm 0mm 0mm,width = 5.8cm,height = 5cm]{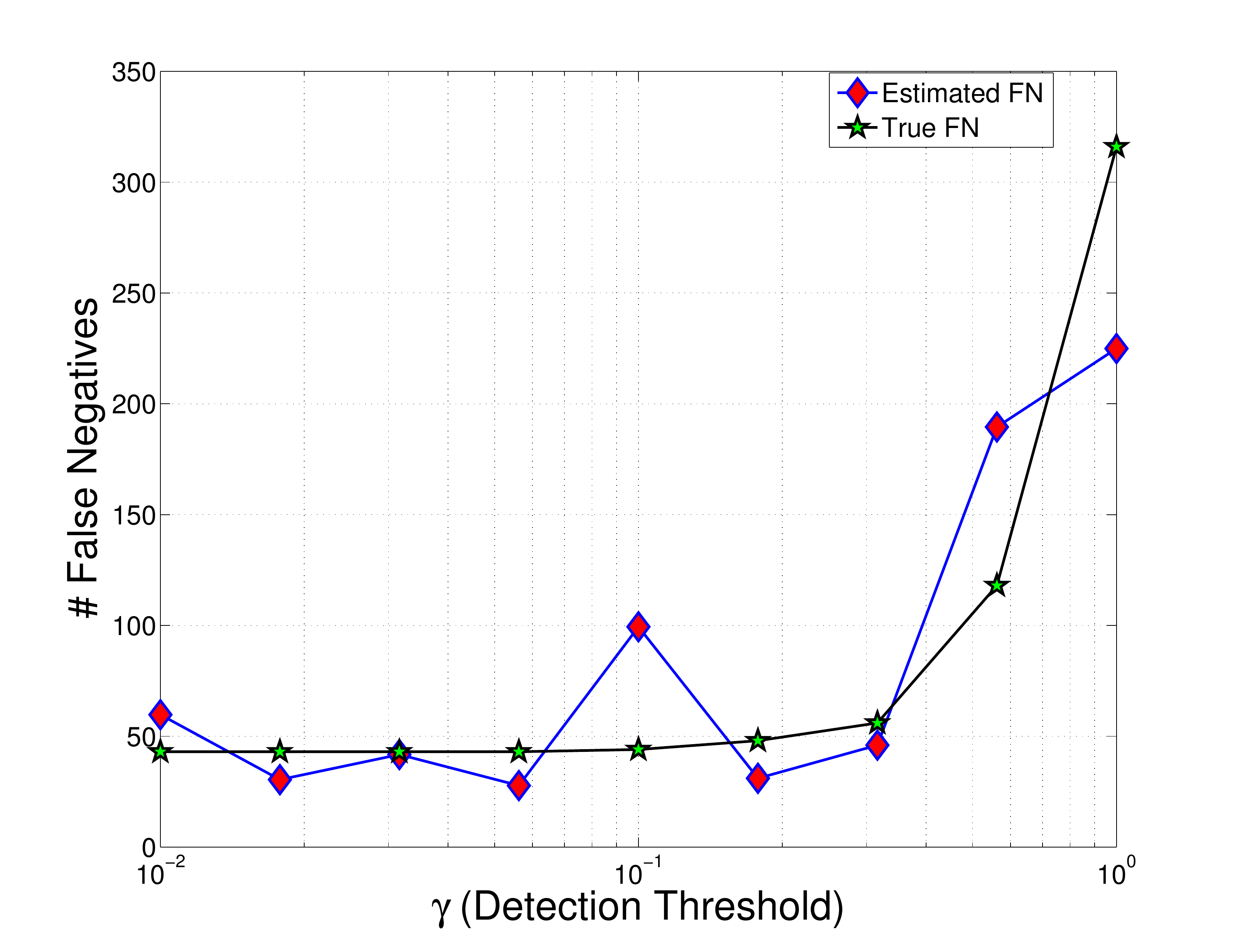}
    \label{fig:ACF_FN}
   }
\subfigure[]{
  \includegraphics[clip = true,trim=10mm 0mm 0mm 0mm,width = 5.8cm,height = 5cm]{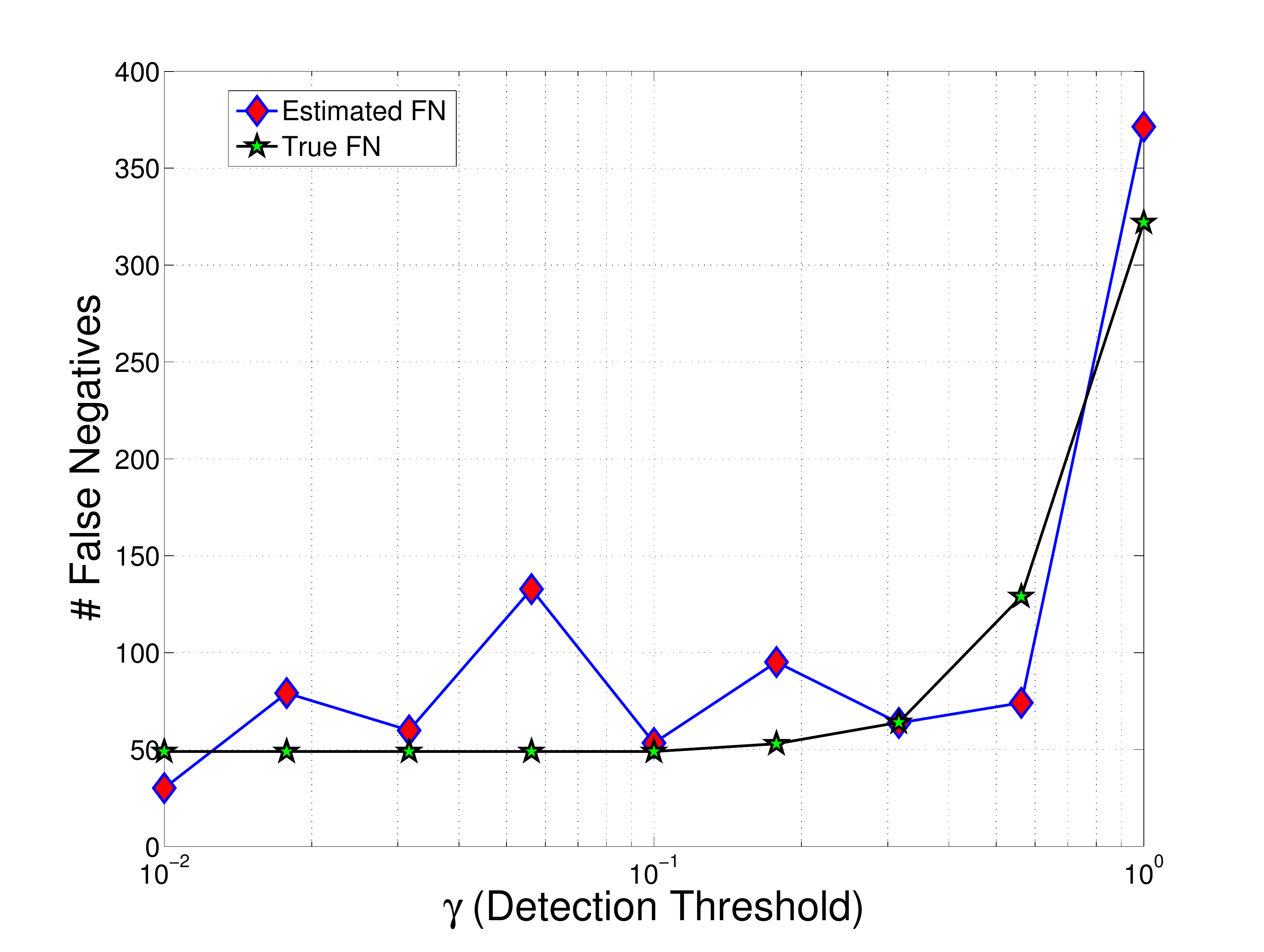}
    \label{fig:ChnFtrs_FN}
 }
 \subfigure[]{
  \includegraphics[clip = true,trim=10mm 0mm 0mm 0mm,width = 5.8cm,height = 5cm]{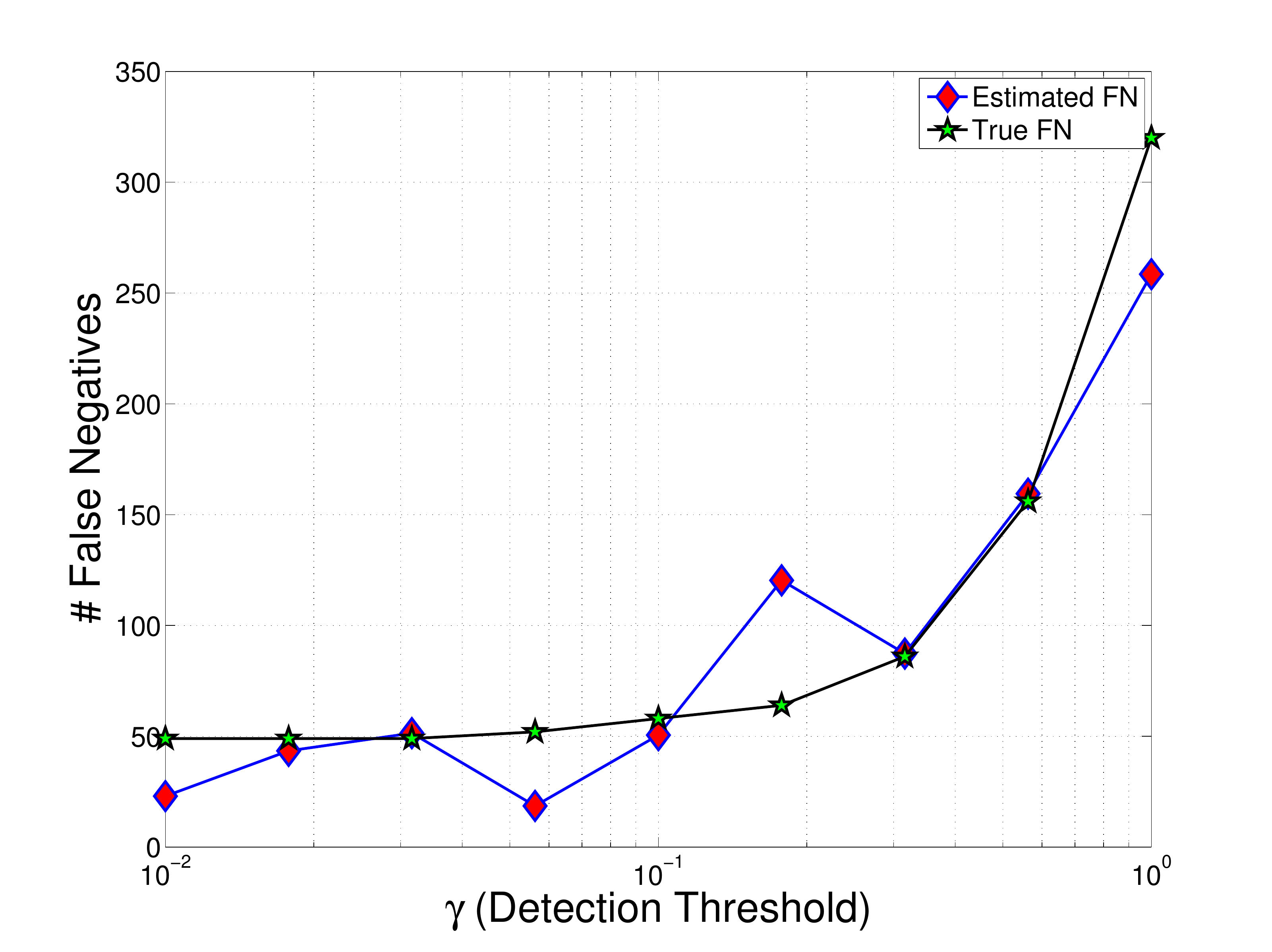}
    \label{fig:Multi_FN}
 }

  \subfigure[ACF Detector]{\includegraphics[clip = true,trim=0mm 0mm 0mm 0mm,width = 5.8cm,height = 5cm]{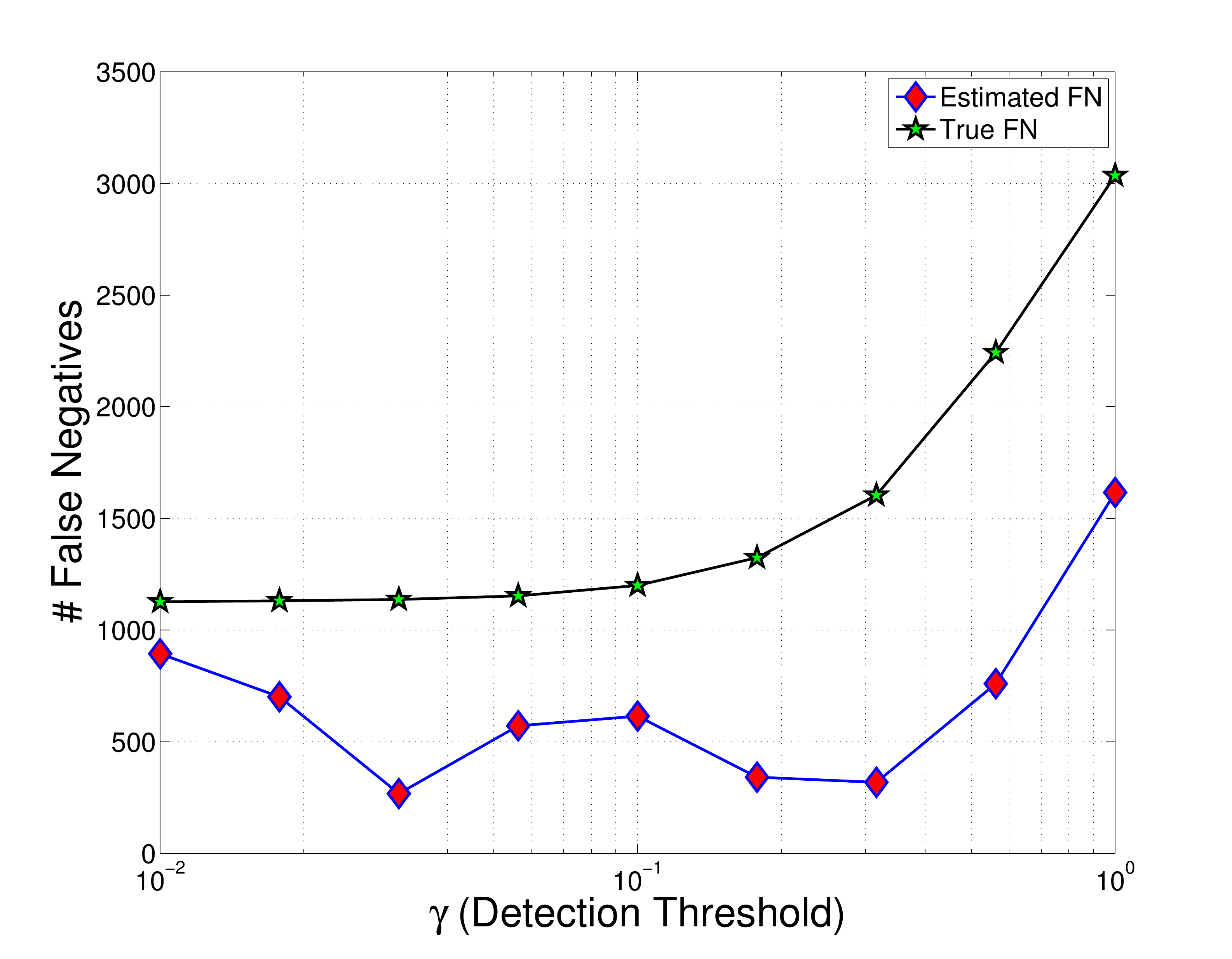}
    \label{fig:ETH_ACF_FN}
   }
\subfigure[ChnFtrs Detector]{
  \includegraphics[clip = true,trim=10mm 0mm 0mm 0mm,width = 5.8cm,height = 5cm]{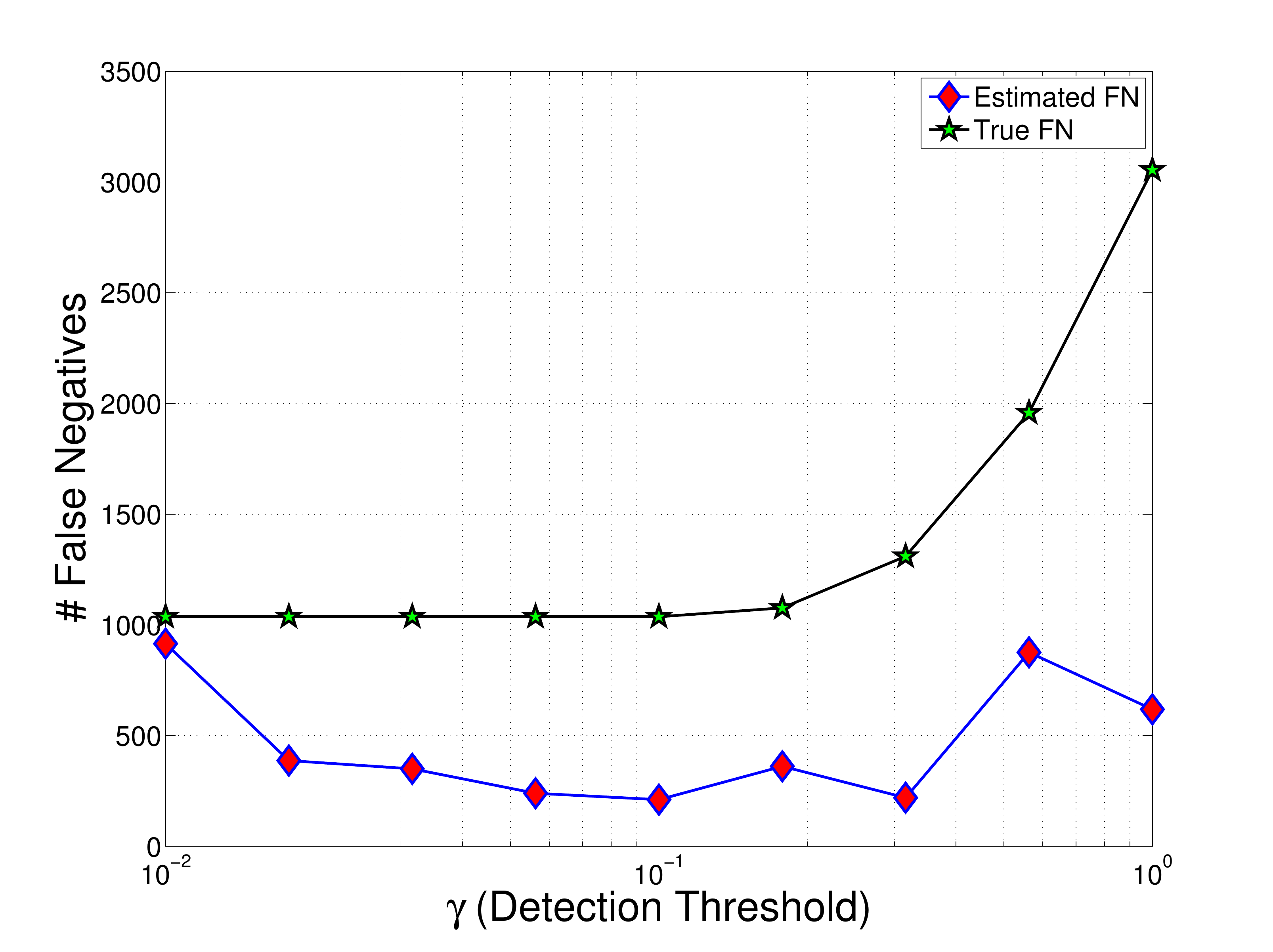}
    \label{fig:ETH_ChnFtrs_FN}
 }
 \subfigure[MultiFtr+CSS Detector]{
  \includegraphics[clip = true,trim=10mm 0mm 0mm 0mm,width = 5.8cm,height = 5cm]{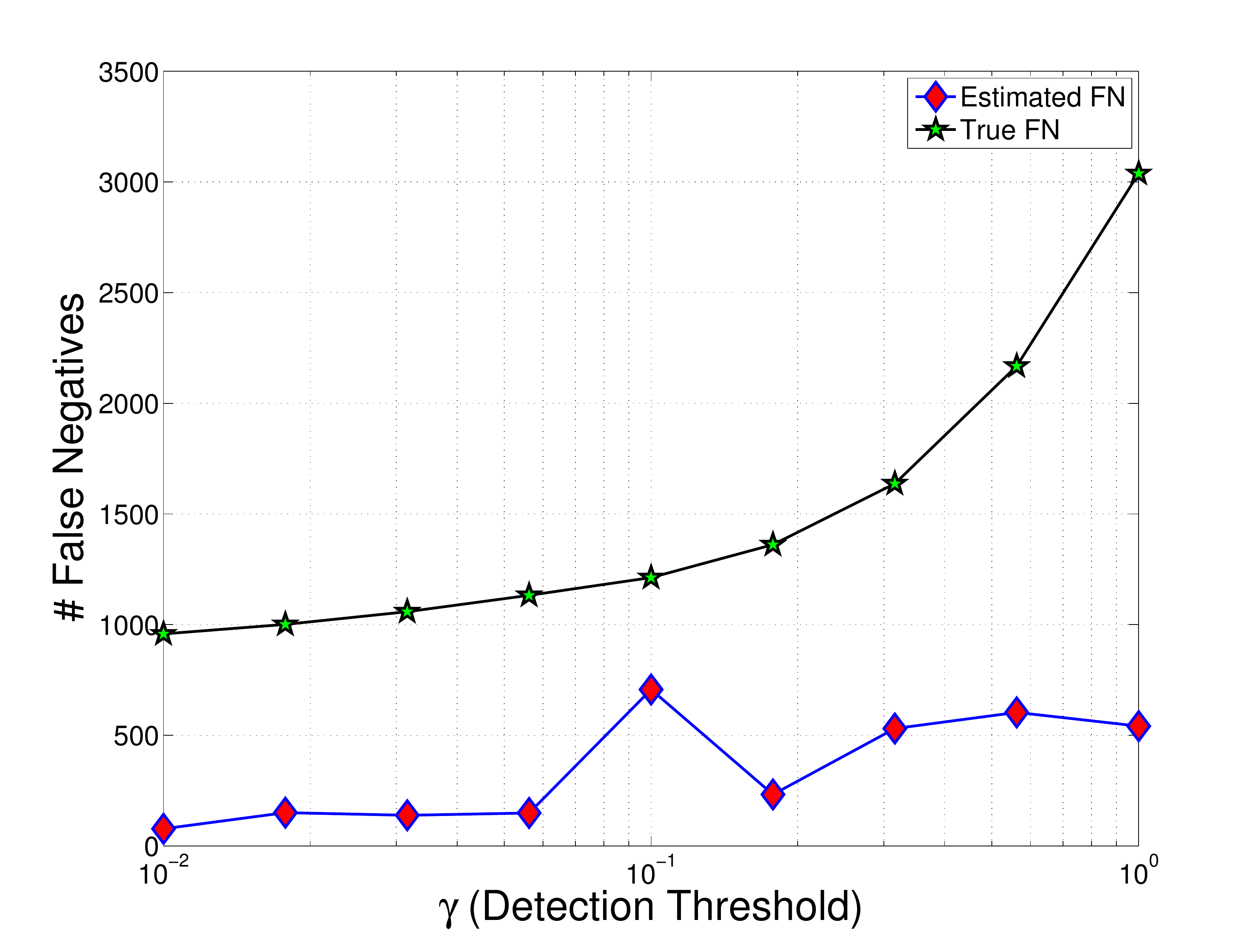}
    \label{fig:ETH_Multi_FN}
 }

     \caption{\small{False negative estimates obtained via pooled testing   on the INRIA \ref{fig:ACF_FN}, \ref{fig:ChnFtrs_FN}, \ref{fig:Multi_FN} \& ETH datasets \ref{fig:ETH_ACF_FN}, \ref{fig:ETH_ChnFtrs_FN}, \ref{fig:ETH_Multi_FN}. The negative bias in the estimates for the ETH dataset is attributed to its complexity as it has multiple pedestrians per frame and constantly varying scale.}}
     \label{fig:ETH_est_recall}
  \vspace{-0.2in}
\end{figure*}

Estimating the number of false negatives is  the second part of evaluating a detector. This is a much harder problem because a negative set is not clearly defined and could be extremely diverse. In obtaining $\overline{D}_\gamma$ for different $\gamma$, the image patches are usually of low resolution (approx. $50\times50$ pixels), and are pooled together before being presented to a human. We found that because of the poor resolution of the images, smaller pool size proved most effective. We used pool size $s=2$ in our experiments and randomly chose two images from the detection complement each time to pool them together. To use pooled testing we use the inverse binomial sampling approach where we fix $n$, the number of objects and make $T$ a variable, from \eqref{eqn:est_p}. So the user is asked to ground truth pooled samples until they find $n$ objects. The choice of $n$ affects the estimate in a way that smaller $n$ gives a quicker estimate but is more noisy. We fixed $n=2$ and found that it was reliable and reasonably quick (in most cases, one can find $2$ images with pedestrians before about $2\%$ of images have been seen). Using \eqref{eqn:est_p} we can get the the ratio $\beta$ of objects within $\overline{D}$, which gives the number of false negatives as $\vert D_{fn}\vert = \beta\times \vert \overline{D}\vert$. Estimated results are shown in fig \ref{fig:ETH_est_recall} for all the detectors and both the datasets. It is observed that the estimated results are poorer in the ETH dataset, this is because the dataset is complex with several pedestrians per frame and of constantly varying scale (the images are taken from a moving camera). Obtaining an accurate $\overline{D}$ for constantly varying object scales is out of the scope of this work and we look towards addressing these issues in the future.

\section{Discussion and Conclusion}
In this paper, we introduce the problem of interactively test-driving object detectors in a user-centric manner, without the need for extensive ground-truthing. We presented the first system that is interactively able to provide estimates of the performance of an object detector without annotations. Although the problem is far from solved, the results shown in this work are promising. Using smart sampling techniques such as summarization and pooled testing, we show that the estimates obtained are close to those obtained using annotations and the effort required is significantly lesser than our baseline techniques. This points towards interactively evaluating classifiers, detectors etc. easily when purchasing them for tailored applications in the future.
\small{
\bibliographystyle{ieee}
\bibliography{ref1,ref2}
}
\end{document}